\documentclass[conference]{IEEEtran}

\usepackage{cite}
\usepackage{amsmath,amssymb,amsfonts}
\usepackage{algorithmic}
\usepackage{graphicx}
\usepackage{textcomp}
\usepackage{xcolor}
\usepackage[utf8]{inputenc}
\usepackage[bookmarks=false]{hyperref}

\hypersetup{
    colorlinks=true,
    linkcolor=.,
    filecolor=.,
    urlcolor=blue,
    citecolor=.
}
\usepackage[normalem]{ulem}  
\usepackage{xspace}

\def\BibTeX{{\rm B\kern-.05em{\sc i\kern-.025em b}\kern-.08em
    T\kern-.1667em\lower.7ex\hbox{E}\kern-.125emX}}

\IEEEoverridecommandlockouts\IEEEpubid{\makebox[\columnwidth]{978-1-7281-0858-2/19/\$31.00~\copyright~2019 IEEE \hfill} \hspace{\columnsep}\makebox[\columnwidth]{ }}

\begin{document}

\title{High-Resolution Road Vehicle Collision Prediction for the City of Montreal}
\author{
\IEEEauthorblockN{Antoine Hébert$^*$, Timothée Guédon$^*$,
Tristan Glatard, Brigitte Jaumard}
Department of Computer Science and Software Engineering \\
Concordia University, Montréal, Québec, Canada\\
$^*$ These authors have contributed equally
}

\maketitle

\begin{abstract}

Road accidents are an important issue of our modern societies, responsible
for millions of deaths and injuries every year in the world. In Quebec only,
in 2018, road accidents are responsible for 359 deaths and 33 thousands of injuries.
In this paper, we show how one can leverage open datasets of a city like
Montreal, Canada, to create high-resolution accident prediction models, using
big data analytics.
Compared to other studies in road accident prediction, we have a much higher
prediction resolution, i.e., our models predict the occurrence of an accident
within an hour, on road segments defined by intersections.
Such models could be used in the context of road accident prevention, but also
to identify key factors that can lead to a road accident, and consequently, help
elaborate new policies.

We tested various machine learning methods to deal with the severe class imbalance inherent
to accident prediction problems. In particular, we implemented the Balanced Random Forest algorithm, a variant
of the Random Forest machine learning algorithm in Apache Spark. 
Interestingly, we found that in our case,
Balanced Random Forest does not perform significantly better than Random Forest.

Experimental results show that 85\% of road vehicle collisions are detected by our model with a false positive rate of 13\%. The examples identified as positive are likely to correspond to high risk situations.
In addition, we identify the most important predictors of vehicle collisions for the area of Montreal: the count of accidents on the same road segment during previous years, the temperature, the day of the year, the hour and the visibility.

\end{abstract}



\section{Introduction}

The World Health Organization describes the road traffic system as the most
complex and the most dangerous system with which people have to deal every
day~\cite{Peden2004}. In the last few years, the number of road traffic deaths in the world climbed, reaching 1.35 million in 2016~\cite{road_safety_report}. More particularly in Quebec, Canada, 359 people were killed in 2018, more than a thousand were seriously injured and tens of thousands have suffered small injuries\cite{saaq}.

Meanwhile, Big Data Analytics has emerged in the last decade as a set of techniques allowing data scientists to extract meaningful information from large amounts of complex and heterogeneous data \cite{Gandomi2015}. In the context of accident prediction, such techniques provide insights on the conditions leading to an increased risk of road accidents, which in return, can be used to develop traffic-related policies and prevention operations. 


\subsection{Open Data}

Governments, states, provinces and municipalities collect and manage data for their internal operations. In the last decade, an open data movement has emerged that encourages governments to make the data they collect available to the public as ``open data".
Open data is defined as ``structured data that is machine-readable, freely shared, used and built on without restrictions"~\cite{opendata101}. Open data should be easily accessible and published under terms that permit re-use and redistribution by anyone and for any purpose.

Open data is made possible by the progress of information technology which allows the sharing of large amounts of data. In 2009, Canada, USA, UK and New Zealand, announced new initiatives towards opening up public information. It is in this spirit that the Government of Canada launched its first-generation of the Open Data Portal in 2011~\cite{opendata101}, giving access to several public datasets. In 2012, the city of Montreal launched its own open data portal.


\subsection{High-Resolution Road Vehicle Collision Prediction}

With the emergence of open data, governments and municipalities are publishing more and more data. At the same time, the recent progresses in Big Data Analytics have facilitated the processing of large data volumes. This makes it possible to build efficient data models for the study of road accidents.

Accident prediction has been extensively studied in the last decade. 
The goal of accident prediction is usually to provide a measure of the risk of accidents at different points in time and space. 
The occurrence of an accident is the label used to train the model, and the proposed model can be used to identify where and when the risk of accident is significantly higher than average in order to take actions to reduce that risk. 
Note that the model cannot be used to predict whether an accident will occur or not. 
Indeed, in order to accurately predict the occurrence of an accident, additional data would be needed: the occurrence of an accident depends on many factors, including driver behavior, that cannot be easily measured. 

Several studies used relatively small datasets and performed accident prediction only on a few selected roads~\cite{Chang2005, Chang2005b, Lin2015, Theofilatos2017}. More recently, other studies performed accident prediction at a larger scale, such as cities or states, using deep
learning\cite{QChen2016, Najjar2017, Yuan2018}.
However, unlike previous studies, they only provide an estimation of the risk of accidents for large areas, i.e., at a coarse spatial resolution. An online article\cite{mediumarticle} presents a study of high resolution road accident prediction in the state of Utah with good performances. This article has inspired us to build a machine-learning model for high-resolution road vehicle collision prediction using public datasets. We used datasets provided by the city of Montreal and the government of Canada as part of their open data initiative. Compared to \cite{mediumarticle}, we have a smaller study area, the island of Montreal, but a much higher prediction resolution. Indeed, the size and precision of our datasets made it possible to predict the occurrence of an accident within an hour on road segments defined by road intersections.

Road vehicle collision prediction can be seen as: 
(1) a regression problem: predicting the risk of accidents, which can be translated into different ways, or 
(2) a binary classification problem: predicting whether an accident will occur. 
We choose to approach it as a classification problem because this simpler approach facilitates the interpretation and comparison of results.
In addition, classification models also provide a measure of probability that can be considered as the risk of an accident.


\subsection{The Data Imbalance Issue}

Like many real-world binary classification problems such as medical diagnosis or fraud prediction, vehicle collision prediction suffers from the data imbalance issue. This issue arises when we are interested in the prediction of a rare event. In this case, the dataset contains much less examples of the class corresponding to the rare event, the positive class. When dealing with severe data imbalance, most machine learning algorithms do not perform well. Indeed, they try to minimize the overall error rate instead of focusing on the detection of the positive class~\cite{Chen2004}.


\subsection{Our Contributions}

In this study, we assembled a dataset containing road vehicle collisions, a dataset describing the Canadian road network, and a dataset containing historical weather information. Using these datasets, we created positive examples, corresponding to the occurrence of a collision, and negative examples, corresponding to the non-occurrence of a collision. For each example, we extracted from the datasets relevant features for accident prediction. Then, we built several prediction models using these examples using various machine learning algorithms. 
We focused on tree-based machine-learning algorithms because they have already proven their effectiveness compared to classical statistical methods \cite{Chang2005, Chang2005b}. In addition, they allow for easier interpretation than deep learning algorithms. We first used the Random Forest algorithm\cite{Breiman2001}. 
We then used the Balanced Random Forest (BRF) algorithm\cite{Chen2004}, a variation of Random Forest specifically designed to better manage data imbalance. 
As BRF was not yet implemented in Apache Spark, we implemented it ourselves. Finally, we considered the XGBoost algorithm\cite{TChen2016}, a gradient tree boosting algorithm which has been used successfully for many machine learning problems and can handle data imbalance\cite{xgboost_doc}.

The contributions of this paper include: 
\begin{itemize}
\item A demonstration of how open datasets can be combined to obtain
meaningful features for road accident prediction,
\item A high spatial and temporal resolution road accident prediction model for the island of Montreal,
\item A comparison of three algorithms dealing with data imbalance in the context of road accident prediction,
\item An implementation of Balanced Random Forest~\cite{Chen2004} in Apache Spark for efficient distributed training.
\end{itemize}

All the source code used is publicly available \href{https://github.com/big-data-lab-team/accident-prediction-montreal}{on Github} under MIT license.

Compared to other studies in accident prediction, our study is original by the size of the datasets used and the spatial resolution of the predictions of our models. 
Previous studies did either use a large dataset (millions of records in total including hundreds of thousands of positive samples \cite{QChen2016}) or predict at a high resolution on one particular road, but no study combines both aspects, which is the hallmark of our study. In terms of prediction resolution, some studies worked on only one road \cite{Chang2005} \cite{Chang2005b} \cite{Lin2015} while some others worked on regions (for example 5km by 5km \cite{QChen2016} or 500m by 500m \cite{Yuan2018}). The road accident dataset we used also covers a wider time range than some studies and is about the maximum time range encountered in the related papers we studied: 7 years \cite{Yuan2018} (against 6 years in our case). For example, other studies have worked on accidents occurring during one year \cite{Chang2005} \cite{Chang2005b} \cite{QChen2016} \cite{Lin2015}. 
In our opinion, predicting at a higher resolution yields more useful results.

The rest of this paper is organized as follows: Section \ref{sec:relatedwork} presents
the related work on accident prediction and on learning with imbalanced data, Section \ref{sec:datasetsintegration} presents the datasets we used and
how we combined them to create positive and negative examples for road
accident prediction, Section \ref{sec:modeldev} presents how we performed feature
engineering, feature selection and hyper-parameter tuning, Section \ref{sec:results}
presents our results and Section \ref{sec:discussion} discusses them.
Conclusions are drawn in the last section.

\section{Related Work}
\label{sec:relatedwork}
\subsection{Road Accident Prediction}
Accident prediction has been extensively studied in the last decades.
Historically, variations of the Poisson regression such as the negative
binomial regression were used to predict the number of accidents that
occurred on a given road segment \cite{Milton1998}. During the last decade,
machine learning algorithms such as decision trees, artificial neural networks
and Bayesian networks have been used successfully to predict road accidents
\cite{Chang2005, Chang2005b, Lin2015, Theofilatos2017}.
Data features usually include information about the road such as number of
lanes, average daily traffic, and road curvature, as well as weather
information such as average precipitation and temperature. 

In 2005,
Chang~\cite{Chang2005} compared the performances of a negative binomial
regression with that of an Artificial Neural Network (ANN) to predict the number
of accidents during a year on road segments of a major freeway in
Taiwan. The dataset contained data from the years 1997 and 1998, which
resulted in 1,338 accidents. The ANN achieved slightly better results than negative 
binomial regression, with
an accuracy of $61.4\%$. On the same dataset, Chang \textit{et al.} \cite{Chang2005b} also used decision trees for accident prediction,
 to get more insights on the important variables for accident
prediction. It appeared that the average daily traffic and the number of
days with precipitation were the most relevant features. The decision tree
reached an accuracy of $52.6\%$. 

Lin \textit{et al.} ~\cite{Lin2015} compared the performances of Frequent Pattern trees\cite{Han2004} with
that of Random Forest for feature selection. They used $k$-nearest-neighbor
and Bayesian networks for real-time accident prediction on a segment of
a highway. Using the mean and sometimes the standard deviation of the weather condition, the visibility, the traffic volume, the traffic speed, and the occupancy measured during the last few minutes their models predict the occurrence of an accident. They obtained
the best results using the Frequent Pattern trees feature selection and achieved
an accuracy of $61.7\%$. It should be noted that they used only a small sample of the
possible negative examples, to deal with data imbalance. 

Theofilatos\cite{Theofilatos2017} also used
real-time data on two urban arterials of the city of Athens to study road
accident likelihood and severity. Random Forest were used for feature
selection and a Bayesian logistic regression for accident likelihood
prediction. The most important features identified were the coefficients of
variation of the flow per lane, the speed, and the occupancy.

In addition, many
studies aim at predicting the severity of an accident using various
information from the accident in order to understand what causes an
accident to be fatal. Chong \textit{et al.} \cite{Chong2005} used decision trees,
neural networks and a hybrid model using a decision tree and a neural
network. They obtained the best performances with the hybrid model which
reached an accuracy of $90\%$ for the prediction of fatal injuries. They
identified that the seat belt usage, the light conditions and the alcohol
usage of the driver are the most important features. Abellán \textit{et al.}
\cite{Abellan2013} also studied traffic accident severity by looking at the
decision rules of a decision tree using a dataset of 1,801 highway
accidents. They found that the type and cause of the accident, the light
condition, the sex of the driver and the weather were the most important
features.

All of these studies use relatively small datasets using data from only a
few years or only a few roads. Indeed, it can be hard to collect all the
necessary information to perform road accident prediction on a larger
scale, and dealing with big datasets is more difficult. However, more
recent studies \cite{QChen2016, Najjar2017, Yuan2018} performed accident prediction at a much larger scale,
usually using deep learning models. Deep learning models can be trained
online so that the whole dataset does not need to stay in memory. This
makes it easier to deal with big datasets.

Chen \textit{et al.} \cite{QChen2016} used human mobility information coming from
mobile phone GPS data and historical accident records to build a model for
real-time prediction of traffic accident risk in areas of 500 by 500
meters. The risk level of an area is defined as the sum of the severity of
accidents that occurred in the area during the hour. Their model achieves
a Root Mean-Square Error (RMSE) of $1.0$ accident severity. They compared the performance of their deep learning
model with the performances of a few classical machine learning
algorithms: Decision Tree, Logistic Regression and Support Vector
Machine (SVM), which all got worse RMSE values of respectively $1.41$,
$1.41$ and $1.73$. We note that they have not tried the Random
Forest algorithm while it usually has good prediction performances. Najjar et
al. \cite{Najjar2017}, trained a convolutional neural network using
historical accident data and satellite images to predict the risk of
accidents on an intersection using the satellite image of the intersection.
Their best model reaches an accuracy of $73\%$. Yuan \textit{et al.} \cite{Yuan2018}
used an ensemble of Convolutional Long Short-Term Memory (LSTM) neural
networks for road accident prediction in the state of Iowa. Each neural
network of the ensemble is predicting on a different spatial zone so that
each neural network learns the patterns corresponding to its zone, which
might be a rural zone with highways or an urban zone. They used a
high-resolution rainfall dataset, a weather dataset, a road network
dataset, a satellite image and the data from traffic cameras. Their model
reaches an RMSE of $0.116$ for the prediction of the number of accidents during
a day in an area of 25 square kilometers.

These more recent studies are particularly interesting because they achieve
good results for the prediction of road accidents in time and space in
larger areas than previous studies which focused on a few roads. But unlike
previous studies, they only provide an estimation of the risk of accidents
for large areas, i.e., at a coarse spatial resolution.
In our study, we decided to focus on urban accidents
occurring in the island of Montreal, a 500-km$^2$ urban area, but with a much higher
prediction resolution. We used a time resolution of one hour and a spatial
resolution defined by the road segments delimited by road intersections. The road
segments used have an average length of 124 meters, and $82\%$ of the road
segments are less than 200 meters long.

Some of these studies define the road accident prediction problem as a 
classification problem, while others define it as a regression problem.
Most of the studies performing classification only report the accuracy
metric which is not well suited for problems with data imbalance such as
road accident prediction\cite{He2009}. The studies performing regression
use different definitions for the risk of accidents, which makes comparisons
difficult.

\subsection{Dealing with Data Imbalance}

Road accident prediction suffers from a data imbalance issue. Indeed, a road
accident is a very rare event so we have much more examples without accident, than examples with accidents available. Machine learning algorithms usually
have difficulty learning from imbalanced datasets \cite{Branco2016}.
There are two main types of approaches to deal with data imbalance. The sampling approaches consist in re-sampling the dataset to make it balanced either by over sampling the
minority class, by under-sampling the majority class or by doing both.
Random under-sampling of the majority class usually performs better than
more advanced methods like SMOTE or NearMiss~\cite{Branco2016}.
The cost-based approach consists in adding weights on the examples. The
negative examples receive a lower weight in order to compensate for their
higher number. These weights are used differently depending on the machine
learning algorithm. 

Chen, Liaw, and Breiman\cite{Chen2004} proposed two methods to deal with class imbalance
when using Random Forest: Weighted Random Forest and Balanced Random Forest.
Weighted Random Forest (WRF) belongs to the class of cost-based approaches. It consists in giving more weight to the minority class when building a tree: during split selection and during 
class prediction of each terminal node. Balanced Random Forest belongs to the class of sampling
approaches. It is similar to Random Forest, but with a
difference during the bootstrapping phase: for each tree of the forest, a random under-sampling of the
majority class is performed in order to obtain a balanced sample. Intuitively,
Balanced Random Forest is an adaptation of random under-sampling of the majority
class making use of the fact that Random Forest is an ensemble method.
While none of the methods is clearly better than the other in terms of predictive
power, BRF has an advantage in terms of training speed because of the under-sampling. Interestingly, Wallace \textit{et al.}~\cite{Wallace2011} present a theoretical analysis of the data
imbalance problem and suggest to use methods similar to Balanced Random Forest.


\section{Datasets Integration}
\label{sec:datasetsintegration}

\subsection{Open Datasets}
\label{sec:datasets}

We used three public datasets\cite{MVC, NRN, HCD} provided by the city of Montreal and the government of Canada: 

\paragraph{\href{http://donnees.ville.montreal.qc.ca/dataset/collisions-routieres}{Montreal Vehicle Collisions}\cite{MVC}}

This dataset, provided by the city of Montreal, contains all the road
collisions reported by the police occurring from 2012 to 2018 on the island
of Montreal. For each accident, the dataset contains the date and
localization of the accident, information on the number of injuries and
deaths, the number of vehicles involved, and information on the road
conditions. The dataset contains 150,000 collisions, among which 134,489
contain the date, the hour and the location of the accident. We used
only these three variables since we do not have other information when no accident happened.
Another dataset with all \href{https://open.canada.ca/data/en/dataset/1eb9eba7-71d1-4b30-9fb1-30cbdab7e63a}{vehicle collisions in Canada} is available but
without the location of the accident, therefore we restrained our
analysis to the city of Montreal.

\paragraph{\href{https://open.canada.ca/data/en/dataset/3d282116-e556-400c-9306-ca1a3cada77f}{National Road Network}\cite{NRN}}

This dataset, provided by the government of Canada, contains the geometry of
all roads in Canada. For each road segment, a few meta-data are given. For
roads in Québec, only the name of the road and the name of the location are provided. The
data was available in various formats, we chose to use the Keyhole Markup
Language, which is a standard of the Open Geospatial Consortium since 2008\cite{kml}, 
This format is based on the Extensible Markup Language (XML), which makes it
easier to read using existing implementations of XML parsers. From this
dataset, we selected the $44,111$ road segments belonging to the island of
Montreal (the dataset is separated into regions and cities).

\paragraph{\href{http://climate.weather.gc.ca/}{Historical Climate Dataset}\cite{HCD}}

This dataset, provided by the government of Canada, contains hourly weather
information measured at different weather stations across Canada. For each
station and every hour, the dataset provides the temperature, the dew point
temperature (a measure of humidity), the humidity percentage,
the wind direction, the wind speed, the visibility, the atmospheric pressure,
the Hmdx index (a measure of felt temperature) and the wind chill 
 (another measure of felt temperature using wind information).
This dataset also contains the observations of atmospheric phenomena such
as snow, fog, rain, etc.

\subsection{Positive and Negative Examples Generation}

The accident prediction problem can be stated as a binary classification
problem, where the positive class is the occurrence of an accident and the
negative class is the non-occurrence of an accident on a given road at a
given date and hour. For each accident, we identified the
corresponding road segment using its GPS coordinates. Such time-road segment pairs are used as
positive examples. For the negative examples, we generated a uniform random sample
of $0.1\%$ of the 2.3 billions possible combinations of time and road segments
in order to obtain 2.3 million examples. We removed from these examples the few
ones corresponding to a collision in the collision dataset in order to obtain
the negative examples.

The identification of the road segments for each collision and the estimation of the weather
information for each road segment made our dataset generation expensive in resources and time. 
We used the big data framework Apache Spark \cite{Zaharia2016} to implement
these dataset combination operations. 
Inspired by the Map Reduce programming model~\cite{mapreduce}, Apache Spark's programming model introduced a new
distributed collection called Resilient Distributed Dataset (RDD), which
provides the ``same optimization as specialized Big Data engines but using it
as libraries" through a unified API. After its release in 2010, Apache Spark
rapidly became the most active open-source project for Big Data~\cite{spark}.
As a consequence, it benefits from a wide community and offers its Application
Programming Interface (API) in the Java, Scala, R and Python programming languages. 

Apache Spark's dataframe API, a collection based on RDDs and optimized for 
structured data processing, is particularly adequate for
combining several datasets. Still, our first implementation had impractical
time and memory space requirements to generate the dataset. Indeed, it was querying
the Historical Climate Data API in real-time with a cache mechanism.
Collecting only the weather stations and hours necessary for our
sample of negative examples resulted in bad performances. We got a
performance increase by first building a Spark dataframe with all the
Historical Climate Data for weather stations around Montreal and then
merging the two datasets. We conducted a detailed analysis of our algorithm
to improve its performances. We notably obtained a good performance
increase by not keeping intermediate results of the road segment
identification for accidents. As opposed to what we initially thought,
recomputing these results was faster than writing and reading them in the
cache. Finally, the identification of the road segment corresponding to
accidents was very memory intensive, we modified this step to be executed
by batches of one month. With these improvements and a few other implementation improvements
including re-partitioning the data frame at key points in our algorithm, we
managed to reduce the processing times to a reasonable time of a few hours.

We also used clusters from Compute Canada to take maximum advantage of the Apache Spark distributed nature for the generation of examples and the hyper-parameter tuning of our models. We started with the Cedar cluster provided by West Grid and we continued with the new B\'eluga cluster provided by Calcul Québec.

To facilitate tests and development, our \href{https://github.com/big-data-lab-team/accident-prediction-montreal/blob/master/src/preprocess.py}{pre-processing program} saves intermediate results to disk in the Parquet format. During later execution of the algorithm, if the intermediate results exists on disk, they will be read instead of being recomputed. This made it possible to quickly test new features and different parameters by recomputing only the required parts of the dataset.

\section{Model Development}
\label{sec:modeldev}

\subsection{Implementation of Balanced Random Forest}

The Balanced Random Forest algorithm was not available in Apache Spark.
An implementation is available in the Python library
imbalanced-learn\cite{imbalance} which implements many algorithms to deal
with data imbalance using an API inspired by scikit-learn\cite{imbalance}, but the size of
our dataset made it impossible for us to use this library. Therefore, we
implemented Balanced Random Forest in Apache Spark.

In the Apache Spark implementation of Random Forest, the bootstrap step is
made before starting to grow any tree. For each sample, an array contains
the number of times it will appear in each tree. When doing sampling with
replacement, values in this array are sampled from a Poisson distribution.
The parameter of the Poisson distribution corresponds to the sub-sampling
rate hyper-parameter of the Random Forest, which specifies the size of the
sample used for training each tree as a fraction of the total size of the
dataset. Indeed, if for example we want each tree to use a sample of the
same size as the whole dataset, the sub-sampling ratio will be set to 1.0,
which is indeed the average number of times a given example will appear in a tree.

To implement Balanced Random Forest, we modified the parameter of
the Poisson distribution to use the class weight multiplied by the
sub-sampling ratio. Hence, a negative sample with a weight
of, say, 0.25 has 4 times less chance to be chosen to appear in a given tree. This
implementation has the advantage that it did not require a big code change
and is easy to test. However, it also has the drawback that users probably
expect linearly correlated weights to be equivalent, which is not the case
in our implementation since multiplying all the weights by $n$ is like multiplying
the sub-sampling ratio by $n$.

To be compatible with other possible use cases, the weights are
actually applied per samples and not per class. This is a choice made by
Apache Spark developers that we respected. To support sample
weights, we create a new Poisson distribution for each sample. To make sure
the random number generator is not reseeded for each sample, we use the
same underlying random number generator for all Poisson distributions, this
also helps reducing the cost of creating a new Poisson distribution object.
Like with other estimators accepting weights, our Balanced Random Forest
implementation reads weights from a weight column in the samples data frame.
We adapted the Python wrapper of the Random Forest classifier to accept and
forward weights to the algorithm in Scala.

\subsection{Feature Engineering}

For each example, we created three types of features: weather features,
features from the road segment, and features from the date and time.

For weather features, we used data from the Historical Climate Dataset (see Section~\ref{sec:datasets}).
To estimate the weather information at the location of the road
segment, we used the mean of the weather information from all the
surrounding weather stations at the date and hour of the example, weighted
by the inverse squared distance between the station and the
road segment. We initially used the inverse of the distance, but we
obtained a small performance improvement when squaring the inverse of
the distance. We tried higher exponents, but the results were not as good.
We used all the continuous weather information provided
by the Historical Climate Dataset. In addition, we created a feature to use
the observations of atmospheric phenomenon provided by the dataset.
To create this feature, we first created a binary variable set to 1 if the
following phenomena are observed during the hour at a given station:
freezing rain, freezing drizzle, snow, snow grains, ice crystals, ice pellets,
ice pellet showers, snow showers, snow pellets, ice fog, blowing snow, freezing
fog. We selected these phenomena because they are likely to increase the risk of
accidents. Then we computed the exponential moving average of this binary
variable over time for each station in order to model the fact that
these phenomena have an impact after they stop being observed and a greater
impact when they are observed for a longer period of time.
We used the same method as for other weather information to get a value for a given 
GPS position from the values of the weather stations.

For the features from the road segments, we were restricted by the limited
metadata provided on the road segments. From the shape of the road segment,
we computed the length of the road segment, and from the name of the
street, we identified the type of road (highway, street, boulevard, etc.).
In addition, road segments are classified into three different levels in
the dataset depending on their importance in the road network: we created a
categorical feature from this information. For these two categorical
features, we encoded them as suggested in The Elements of Statistical
Learning~\cite{elementsofstat} in Section 9.2.4. Indeed, instead of using one-hot
encoding which would create an exponential number of possible splits, we indexed the
categorical variable ordered by the proportion of the examples belonging to
the given category, which are positive samples. This encoding guarantees 
optimal splits on these categorical variables. Lastly, we added a
feature giving the number of accidents that occurred previously on this
road segment.

For the date features, we took the day of the year, the hour of the day,
and the day of the week. We decided to make the features ``day of the year"
and ``hour of the day" cyclic. Cyclic features are used when the extreme values
of a variable have a similar meaning. For example, the value 23 and 0 for
the variable hour of the day have a close meaning because there is only one
hour difference between these two values. Cyclical encoding allows this fact to be expressed.
With cyclical encoding, we compute two features, the first one is
the cosine of the original feature scaled between 0 and 2$\pi$, and the second
one is the sine of the original feature scaled between 0 and 2$\pi$.
In addition to these basic date features, \href{https://github.com/big-data-lab-team/accident-prediction-montreal/blob/master/src/solar\_features.py}{we computed} an approximation
of the solar elevation using the hour of the day, the day of the year
and the GPS coordinates. The solar elevation is the angle 
between the horizon and the sun. Note that it is of interest, because it is
linked to the luminosity which is relevant for road accident prediction.

\subsection{Identifying the most Important Features}

Random Forest measures feature importance by computing the total
decrease in impurity of all splits that use the feature, weighted by the
number of samples. This feature importance measure is not perfect for
interpretability since it is biased toward non-correlated variables, but it
helps selecting the most useful features for the prediction. 
Random Forest usually performs better when irrelevant features are removed.
Therefore, we removed the features wind direction, wind speed, dew
point temperature, wind chill, hmdx index and day of month which had a
much lower feature importance. This improved the performances of the model.

\subsection{Hyper-Parameter Tuning}

To determine the optimized hyper parameters, we first performed
automatic hyper-parameter tuning by performing a grid search 
with cross-validation. Because the processing times on the whole dataset would
have been too high, we took a small sample of the dataset. Still, we could
not test many parameter combinations using this method.

Once we got a first result with grid search we continued manually by
following a plan, do, check, adjust method. We plotted the precision-recall and ROC curves on the test and training set to understand how the performances of our model could be improved. These curves are obtained by computing the precision, the recall and the false positive rate metrics when varying the threshold used to classify an example as positive. Most classification algorithms provide a measure of the confidence with which an example belongs to a class. 
We can reduce the threshold on the confidence beyond which we classify the example as positive in order to obtain a higher recall but a lower precision and a higher false positive rate.
In order to obtain a general measure of the performances of a classifier
at all thresholds, we can use the area under the ROC curve. The area under the Precision-Recall curve, however, should not be used\cite{flach2015precision}

Interestingly, despite using many trees, our Random Forest classifiers
tended to over-fit very quickly as soon as the maximum depth parameter went
above 18. We eventually used only 100 trees, because adding more trees did
not increase performances. We have not tried more than 200 trees, maybe
many more trees would have been necessary to increase the maximum depth
without over-fitting, but then the memory requirement would become unreasonable.
Our final parametrization used a total of 550 gigabytes of memory per training of the Balanced Random Forest model on the cluster.


\section{Results}
\label{sec:results}


\subsection{Balanced Random Forest Performances}

To test our implementation of Balanced Random Forest (BRF) in Apache Spark, we performed an experiment on an imbalanced dataset provided by the imbalanced-learn library.
We chose to use the mammography dataset\cite{Woods1993} which is a small dataset with 11,183 instances and 6 features. It has an imbalance ratio of 42, i.e., there are 42 times more negative samples than positive samples. We compared the performances obtained with the implementation of BRF in the library imbalanced-learn with those obtained with our implementation of BRF in Apache Spark. We also compared these performances with the performances obtained with both implementations of the classical Random Forest algorithm.
Results are summarized in Table~\ref{table:test_brf_results}. We observe that we obtain similar results with both implementations of BRF.

\begin{table}[htbp]
\caption{Comparison of our BRF implementation with imbalanced-learn}
\begin{center}
\begin{tabular}{|l|r|r|}
\cline{2-2}
\multicolumn{1}{c|}{} &  Area under ROC \\
\hline
imbalanced-learn RF  &   0.932 \\
Spark RF             &   0.951 \\
imbalanced-learn BRF &   0.956 \\
Spark BRF            &   0.960 \\
\hline
\end{tabular}
\label{table:test_brf_results}
\end{center}
\end{table}

\begin{figure}[htbp]
\centerline{\includegraphics[height=6cm, keepaspectratio]{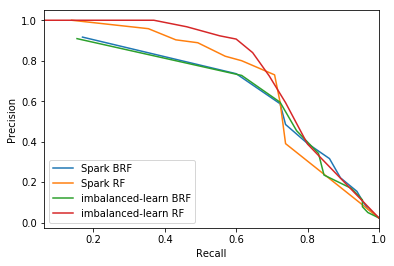}}
\caption{Comparison of implementations: Precision-recall curves}
\label{fig:test-brf-precision-recall}
\end{figure}

Figure~\ref{fig:test-brf-precision-recall} shows the precision-recall curves obtained with both implementations of the Balanced Random Forest (BRF) and Random Forest (RF) algorithms on the mammography dataset. We can see that, with a low recall, BRF implementations perform worse, and with a high recall, all the models have similar performances except the Random Forest model from Apache Spark which has a lower precision.

\begin{figure}[htbp]
\centerline{\includegraphics[height=6cm, keepaspectratio]{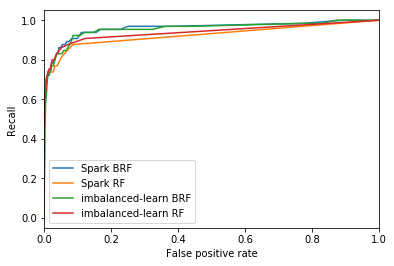}}
\caption{Comparison of implementations: ROC curves}
\label{fig:test-brf-roc}
\end{figure}

Figure~\ref{fig:test-brf-roc} shows the Receiver operating characteristic (ROC) curves obtained with both implementations of the Balanced Random Forest (BRF) and Random Forest (RF) algorithms.

\subsection{Vehicle Collision Prediction}

Results were obtained by training the algorithms on the whole
dataset of positive samples and with a sub-sample of 0.1\% of the 2
billion possible negative examples. This corresponds to a total of 2.3
million examples with a data imbalance reduced to a factor of 17. To
evaluate our models, we used a test set containing the last two years of our
dataset. The model was trained on the 4 previous years and used only data
from these years. For instance, the ``count\_accident" feature contains only
the count of accidents occurring from 2012 to 2016 on the road segment.
In addition to the three models built using tree-based machine learning algorithms,
we created a simple baseline model. This model is very basic in the sense that it uses
only the count of accidents of the road segment.
The probability of accidents given 
by this model for an example whose road segment has a count of accidents of $n$, is the percentage
of positive examples among the examples with a count of accidents higher than $n$.

Table~\ref{table:summary} presents the results obtained on the test set with the classical
Random Forest algorithm with further under-sampling (RF), with the Balanced Random Forest algorithm (BRF), with the XGBoost algorithm (XGB), and with the baseline model (base). The values of the hyper-parameters we used and more details about the results are available on the \href{https://github.com/big-data-lab-team/accident-prediction-montreal/tree/master/results}{Github repository of the project}.

\begin{table}[htbp]
\caption{Result Summary}
\begin{center}
\begin{tabular}{|l|r|r|r|r|}
\cline{2-5}
\multicolumn{1}{c|}{}    &    BRF &    RF  &    XGB & base \\
\hline
Area under the ROC curve &  0.916 &  0.918 &  0.909 & 0.874 \\
\hline
\end{tabular}
\label{table:summary}
\end{center}
\end{table}
As we can see, the three machine learning models obtain similar performances
and perform much better than the baseline model. The XGBoost
model has slightly worse performances than the two others.

\begin{figure}[htbp]
\centerline{\includegraphics[height=6cm, keepaspectratio]{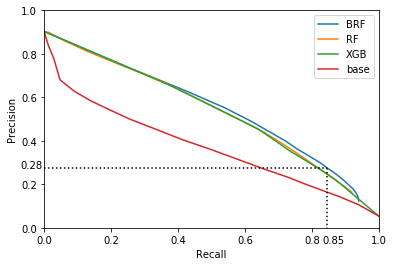}}
\caption{Vehicle Collision Prediction: Precision-recall curves}
\label{fig:precision-recall}
\end{figure}

Figure~\ref{fig:precision-recall} shows the precision-recall curves of the three models.

\begin{figure}[htbp]
\centerline{\includegraphics[height=6cm, keepaspectratio]{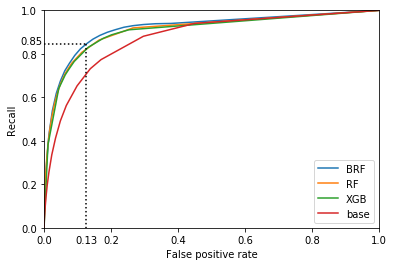}}
\caption{Vehicle Collision Prediction: ROC curves}
\label{fig:roc}
\end{figure}

Figure~\ref{fig:roc} shows the Receiver operating characteristic (ROC) curves of the three models.

\begin{figure}[htbp]
\centerline{\includegraphics[height=6cm, keepaspectratio]{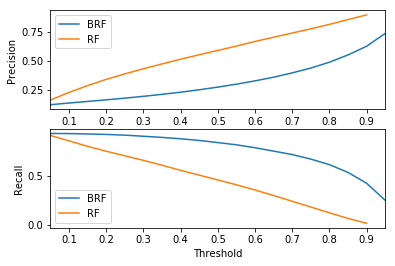}}
\caption{Vehicle Collision Prediction: Precision and Recall as a Function of the Threshold Values}
\label{fig:pr-threshold}
\end{figure}

Figure~\ref{fig:pr-threshold} shows the precision and the recall as a function of the threshold values for BRF and RF algorithms.
It shows that despite BRF and RF having similar results on the PR and ROC curves, they have different behaviors. For an identical threshold value, BRF has a higher recall but a lower precision than RF.

As we can see, the Balanced Random Forest model surprisingly does not perform better than the other models.
It achieves a recall of $85\%$ with a precision of $28\%$, and a false positive rate (FPR) of $13\%$ on the test set.


\subsection{Vehicle Collision Feature Importance}

\begin{figure}[htbp]
\centerline{\includegraphics[height=7cm, keepaspectratio]{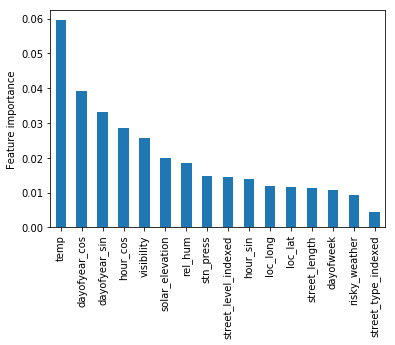}}
\caption{Feature importance computed by the Balanced Random Forest excluding the accident count feature.}
\label{feature importances}
\end{figure}

With a
feature importance of $67\%$, the number of accidents which occurred on the
road segment during the previous years is clearly the most useful feature. 
This shows that accidents are concentrated on specific roads.
Figure~\ref{feature importances} presents the
importance of the other features as reported by the Balanced Random Forest
algorithm. As we can see, the next most important feature is the temperature. 
Then, the day of the year, the cosine of the hour of the day, which separates day from night,
and the visibility follow. The solar elevation and the humidity are 
the following features of importance. The remaining features have almost
the same importance, except the street type which
is significantly less important.

We believe that the road features like the street length, the street level and the street type have a lower importance because the accident count already provides a lot of information on the dangerousness of a road segment. Surprisingly, the risky weather feature is one of the least important ones. 
This suggests that our definition of risky weather may need to be revisited.

As compared to the count of accidents, the other features seem to have almost no
importance, however the performance of the model decreases significantly if we
remove one of them. 

\section{Discussion}
\label{sec:discussion}

With areas under the ROC curve of more than $90\%$, the performances of our models are good.
However, they mostly rely on the count of previous accidents on the road segment as we can see from the feature importance of the accident count feature and the performance of the base model.
This is not an issue for accident prediction, but it does not help to 
understand why these roads are particularly dangerous. 
We believe that this feature is even more useful because we do not have information
about the average traffic volume for each road. 
Therefore, this feature does
not only inform the machine learning algorithm about the dangerousness of a road
segment but also indirectly about the number of vehicles using this road.
Nonetheless, the performance of our models does not only rely on this feature.
As we can see from the curves, the performances of our models are significantly
better than those of the base model that exclusively relies on the count of accidents.

\subsection{Test of our Implementation of BRF on the Mammography Dataset}
As expected, we obtained similar results to the imbalanced-learn library with our implementation of the BRF algorithm. The precision-recall curve shows that the BRF algorithm had a better precision with high recall values, but a much lower precision with low recall values. For medical diagnosis and road vehicle collision prediction, we usually prefer to have a higher recall with a lower precision, so BRF is more suitable for these use cases.

\subsection{Comparison of the Different Models for Road Vehicle Collisions Prediction}
For the road vehicle collision prediction, the Balance Random Forest algorithm obtained slightly better results than the classical Random Forest algorithm. However, the gain in prediction performance is very small. 
We believe this is caused by the fact that negative examples are not so different from each other and the information they contain is well captured by a single random sub-sample. We observe that the BRF algorithm achieved better performances than Random Forest with high recall values. With lower recall values, both Random Forest algorithms had similar performances. The XGBoost algorithm obtained worse results than the two other algorithms. However, it is still interesting because it was much faster to train than Random Forest algorithms. This made the hyper-parameter tuning of the XGBoost algorithm easier and much faster.

\subsection{Real-world Performances of our Road Vehicle Collision Prediction Model}
As stated previously, the accuracy measure is not a good metric for road accident
prediction. Indeed, since most examples belong to the negative class, the model which obtains
the best accuracy is usually the one with the lowest false positive rate. But for rare event
prediction, we usually want a model with a high recall even if it implies a higher false
positive rate. This is especially true in accident prediction, because false positives can correspond to high-risk situations that we probably want to detect too. For these reasons, we decided not to use the accuracy measure. Instead
we used the precision-recall curve to compare the performances of our models. However, we
should be careful when using the precision measure on a dataset using a sample of the possible
negative examples like it is usually the case in accident prediction. Indeed, the precision
computed on the test set does not correspond to the precision we would obtain in production.
If the sample of negative examples is representative of the population
in production, the model will achieve the same false positive rate.
Because we used a sample of the possible negative examples but all the
positive examples in the test set, there will be more cases of false positive in production for the same number of positives. As a consequence, the precision will be much lower.

Since we know the proportion of positive examples in the real world, if we assume that the
sample of negative examples is representative of the population in production, we can provide
an estimation of the precision that the model could achieve. There are on average $22,414$ collisions each year and during a year there are a total of $386,412,360$ combinations of hour and road segments. Therefore, in the real-world approximately $0.0058\%$ of examples are positive. With a recall of $85\%$, approximately $0.00493\%$ of examples are true positives and $0.00087\%$ are false negatives. With a false positive rate of $13\%$, approximately $12.99925\%$ of examples are false positives and $86.99495\%$ are true negatives. Therefore, with the real world distribution, our model would likely obtain a precision of
$0.04\%$. If the goal of our model was to actually predict accidents, this would not be a satisfying precision, but the real goal of accident prediction is to identify when and where the risk of accidents is significantly higher than average in order to take measures. With this precision, the probability of a collision to occur is 6 times higher than average for examples detected as positive. By varying the threshold used by the model, we can choose when to take actions.

\subsection{Reproducibility of the study} 

The results from this study can be reproduced using the \href{https://github.com/big-data-lab-team/accident-prediction-montreal}{Github repository} of the project. The \href{https://github.com/big-data-lab-team/accident-prediction-montreal/blob/master/README.md }{'readme'} file provides more information on how to create training examples from the datasets and how to train the models. All the figures can be reproduced with the \href{https://github.com/big-data-lab-team/accident-prediction-montreal/tree/master/notebooks}{Jupyter notebooks} available on the same repository.

The National Road Network and the Historical Climate datasets used in this study are open datasets from the government of Canada. 
One can potentially reproduce the study for any other Canadian city as long as the the city provides open data on vehicle collisions including the date, time and localization of such collisions in sufficient amount. 
For example, the city of Toronto seems to be a good candidate with 11 years of vehicle collisions open data available through the \href{http://data.torontopolice.on.ca/datasets/automobile}{``Automobile"} dataset provided by the Toronto Police Service. 
The latter dataset contains the date, time and localization of the accidents. National road network information and historical climate information tends to be easily found for many countries which would allow this study to also be reproducible in other countries. For example historical climate information for the United States can be found in the \href{https://data.ess-dive.lbl.gov/view/doi:10.3334/CDIAC/CLI.NDP019}{U.S. Historical Climatology Network} dataset and road network information seems to be available in the \href{https://catalog.data.gov/dataset/usgs-national-transportation-dataset-ntd-downloadable-data-collectionde7d2}{USGS National Transportation Dataset}.


\subsection{Future Work}

We believe that a better performance could be reached by adding more features
from other datasets. For the city of Montreal, we identified two
particularly interesting datasets: a dataset with the location and dates of
construction work on roads, and a dataset with the population density.
In addition, Transport Qu\'ebec gives access to cameras monitoring the main
roads of Montreal. The videos from these cameras could be useful to get an
estimation of the traffic in the roads of the island.
These datasets could be used to improve prediction performances.
However, this type of dataset might not be available for other geographical areas.
The current model use datasets that can easily be made available for most cities.

The most important feature is the number of accidents which happened during
the previous year. While this feature helps a lot to reach useful prediction
performances, it does not help in understanding the characteristics of a
road segment which makes it dangerous. A human analysis of these
particularly risky road segments could detect patterns that could help to
take measure to reduce the number of accidents in Montreal. This can also
be useful to improve our current accident prediction model, if the detected
patterns can be used by merging other datasets.

Lastly, it would be interesting to analyze why BRF did not perform better for this problem in order to understand under which conditions it helps to deal with data imbalance.

\section{Conclusions}

In this study, we conducted an analysis of road vehicle collisions in the
city of Montreal using open data provided by Montreal city and the Government of Canada. 
Using three different datasets, we built road vehicle collision
prediction models using tree-based algorithms. Our best model can predict $85\%$ of road accidents in the area of Montreal with a false positive rate of $13\%$.
Our models predict the occurrence of a collision at high space resolution and hourly precision. In other words, it
means our models can be used to identify the most dangerous road
segments every hour, in order to take actions to reduce the risk of accidents. 
Moreover, we believe that our work
can easily be reproduced for other cities under the condition that similar 
datasets are available. One can freely use our source code on Github for reference.
Finally, our study shows that open data initiatives are useful to society because they make it possible to study critical issues like road
accidents.


\section*{Acknowledgment}

The authors would like to acknowledge Compute Canada for providing access to the computation clusters, as well as WestGrid and Calcul Québec, Compute Canada's regional partners for the clusters used.

\bibliographystyle{IEEEtran}
\IEEEtriggeratref{15}
\bibliography{IEEEabrv,Biblio/biblio}

\end{document}